\documentclass[letterpaper]{article} % DO NOT CHANGE THIS
\usepackage{aaai24}
\usepackage{times}  % DO NOT CHANGE THIS
\usepackage{helvet}  % DO NOT CHANGE THIS
\usepackage{courier}  % DO NOT CHANGE THIS
\usepackage[hyphens]{url}  % DO NOT CHANGE THIS
\usepackage{graphicx} % DO NOT CHANGE THIS
\usepackage{natbib}  % DO NOT CHANGE THIS AND DO NOT ADD ANY OPTIONS TO IT
\usepackage{caption} % DO NOT CHANGE THIS AND DO NOT ADD ANY OPTIONS TO IT
\usepackage{algorithm}
\usepackage{algorithmic}

\usepackage{xspace}
\usepackage{paralist}
\usepackage{amsmath}
\usepackage{amssymb}
\usepackage{tabularx}
\usepackage{booktabs} % For formal tables
\usepackage{makecell}
\usepackage{multirow}
\usepackage{arydshln}

\usepackage[utf8]{inputenc}
\usepackage[T1]{fontenc}
\usepackage{CJKutf8}
\usepackage{xcolor}
\usepackage{colortbl}

\usepackage{accents}
\usepackage{newfloat}
\usepackage{listings}
\DeclareCaptionStyle{ruled}{labelfont=normalfont,labelsep=colon,strut=off} % DO NOT CHANGE THIS
\floatstyle{ruled}
\newfloat{listing}{tb}{lst}{}
\floatname{listing}{Listing}
\newcommand{\idiomkb}{\textsc{IdiomKB}\xspace}

\newcommand{\ie}{\textit{i.e.}\xspace}

\title{\textit{Translate Meanings, Not Just Words}:\\ 
IdiomKB's Role in Optimizing Idiomatic Translation with Language Models}

\author{
     Shuang Li\textsuperscript{\rm 1},
     Jiangjie Chen\textsuperscript{\rm 1}\thanks{Corresponding authors. Yanghua Xiao is also a member of Research Group of Computational and AI Communication at Institute for Global Communications and Integrated Media, Fudan University.},
     Siyu Yuan\textsuperscript{\rm 2},
     Xinyi Wu\textsuperscript{\rm 1},
     Hao Yang\textsuperscript{\rm 3},
     Shimin Tao\textsuperscript{\rm 3},
     Yanghua Xiao\textsuperscript{\rm 1,4}\footnotemark[1]
}
\affiliations{
    \textsuperscript{\rm 1}Shanghai Key Laboratory of Data Science, School of Computer Science, Fudan University\\
    \textsuperscript{\rm 2}School of Data Science, Fudan University \\
    \textsuperscript{\rm 3}Huawei Translation Services Center\\
    \textsuperscript{\rm 4}Fudan-Aishu Cognitive Intelligence Joint Research Center\\
    \text{\{lishuang18, jjchen19, wuxinyi20, shawyh\}@fudan.edu.cn}, \text{syyuan21@m.fudan.edu.cn}, \\
    \text{\{yanghao30, taoshimin\}@huawei.com}
}
\usepackage{bibentry}
\begin{document}
\begin{CJK}{UTF8}{gbsn}
\maketitle

\begin{abstract}

To translate well, machine translation (MT) systems and general-purposed language models (LMs) need a deep understanding of both source and target languages and cultures. 
Therefore, idioms, with their non-compositional nature, pose particular challenges for Transformer-based systems, as literal translations often miss the intended meaning.
Traditional methods, which replace idioms using existing knowledge bases (KBs), often lack scale and context-awareness.
Addressing these challenges, our approach prioritizes context-awareness and scalability, allowing for offline storage of idioms in a manageable KB size. 
This ensures efficient serving with smaller models and provides a more comprehensive understanding of idiomatic expressions.
We introduce a multilingual idiom KB (\idiomkb) developed using large LMs to address this. 
This KB facilitates better translation by smaller models, such as BLOOMZ (7.1B), Alpaca (7B), and InstructGPT (6.7B), by retrieving idioms' figurative meanings. 
We present a novel, GPT-4-powered metric for human-aligned evaluation, demonstrating that \idiomkb considerably boosts model performance. 
Human evaluations further validate our KB's quality.\footnote{Code and resources can be found at \url{https://github.com/lishuang-w/IdiomKB}.}

\end{abstract}
\section{Introduction}
\label{sec:intro}

Idioms are non-compositional expressions whose \textit{figurative meanings} deviate from the meanings of the constituent words (\textit{literal meanings})~\cite{bobrow1973catching,swinney1979access, salton2018exploring,fadaee-etal-2018-examining}.
For example, the figurative meaning of the idiom ``bite the bullet'' is ``to endure a painful situation'', deviating from the literal meanings of the constituent words ``bite'' and ``bullet''.
Given the diverse array of idioms across various cultures and languages, appropriately translating texts that contain idioms (\textit{idiomatic texts}) has become an important research problem~\cite{strakvsiene2009analysis, tzou2017does, shao-etal-2018-evaluating,10.1162/tacl_a_00572}.

\begin{figure}[t]
    \centering
    \includegraphics[width=\linewidth]{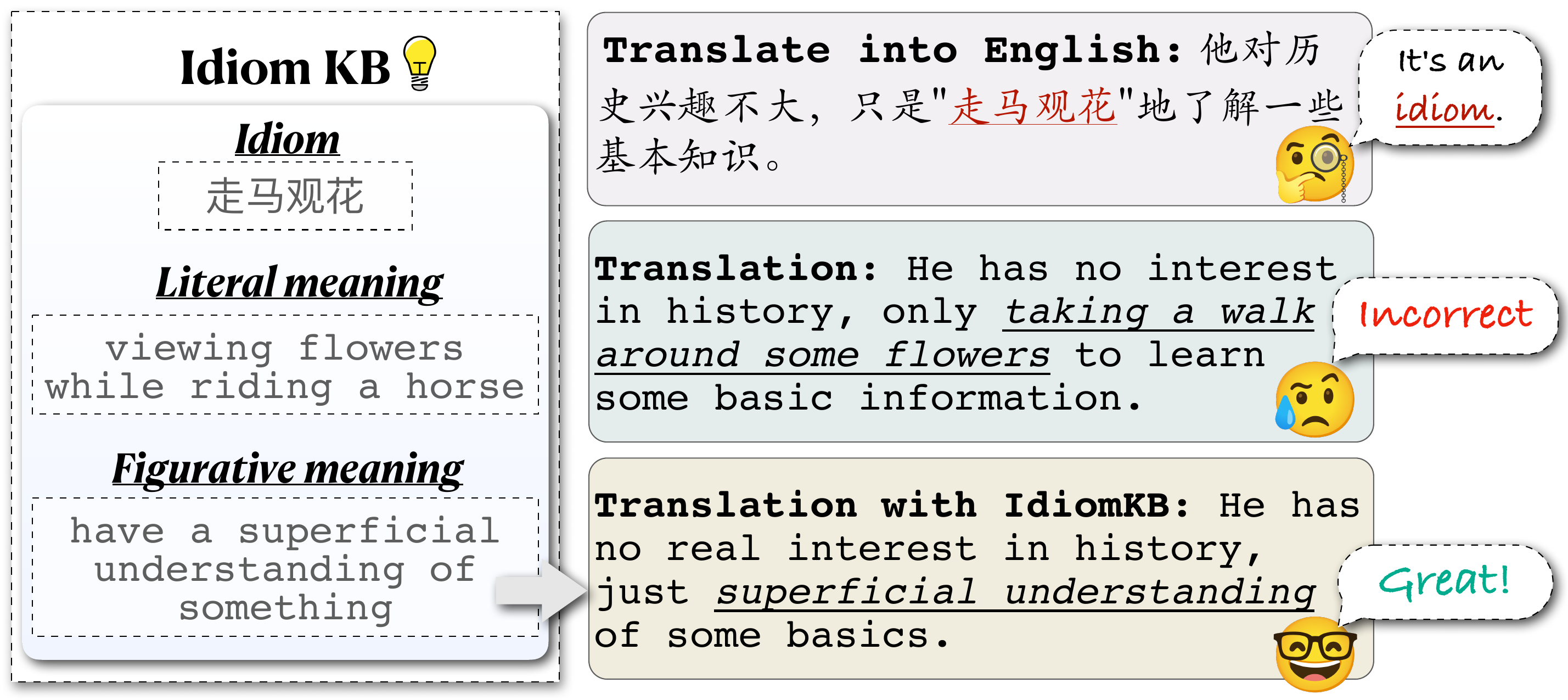}
    \caption{An example of idiomatic translation from Chinese to English. Current machine translation systems incorrectly translate idiomatic texts based on the \textit{literal meaning} of the idiom, resulting in unsatisfactory translation. Incorporating \textit{figurative meaning} from an idiom knowledge base (\idiomkb) improves the translation performance.}
    \label{fig:front}
\end{figure}

However, due to the non-compositionality of idioms, idiomatic translation poses a significant challenge for current machine translation (MT) systems and general-purposed language models (LMs).
Traditional phrase-based statistical machine translation systems do not give special consideration to idioms, resulting in low-quality translations~\cite{salton-etal-2014-empirical, Manojlovic2017IdiomsIS}.
Transformer-based MT models~\cite{vaswani2017attention,dankers-etal-2022-transformer} usually treat idioms as compositional expressions, leading to literal translation errors.
As shown in Figure~\ref{fig:front}, they usually translate idiomatic texts based on the literal meaning, failing to convey the intended information.
One way to address the idiomatic translation problem would be scaling up the model size and training data, \ie, large language models (LLMs)~\cite{ouyang2022training,openai2022chatgpt}, where various strong abilities emerge~\cite{wei2022emergent}.
However, deploying models of such sizes for offline scenarios or real-time responses is costly and demanding.
Therefore, the research question arises: \textit{Can we enable smaller or specialized models to do idiomatic translation better?}

To tackle this challenge, one intuitive solution is to utilize the figurative meanings of the idioms, which are equivalent to semantically literal expressions, as support for translation.
Recent research in linguistics and education has leveraged this insight to develop idiom knowledge bases (KBs), which have been employed to evaluate~\cite{Cucchiarini2020LearningLI,Wang2021} and assist~\cite{jiang-etal-2018-chengyu, tang2022petci} second language learners’ comprehension of idioms and idiomatic texts.
Some studies leverage idiom KBs as valuable transition aids for models to accurately infer idiomatic figurative meanings without dealing with non-compositional expressions~\cite{salton-etal-2014-evaluation, Modh2021UsingIF}.
However, obtaining these meanings from dictionaries or manual annotation is time-consuming and suffers from idiom coverage issues.
As a result, existing idiom KBs are relatively small in scale and notably lack multilingual meanings.
This hinders translations since the idiom in the source sentence may not be included in the KB, or the figurative meaning in the target language could be absent.

In this paper, we propose \idiomkb, a machine-generated KB consisting of idioms and their multilingual meanings, to facilitate the translation of idiomatic texts for LM-based MT systems.
To circumvent the need for labor-intensive and expensive human efforts, we adopt the idea of symbolic knowledge distillation~\cite{west-etal-2022-symbolic, yuan-etal-2023-distilling, bhagavatula-etal-2023-i2d2} and employ LLMs to distill multilingual figurative meanings of the idioms based on their powerful generation ability.
Based on the human evaluation, our \idiomkb exhibits high quality, with an average score of 2.92 out of 3.
Furthermore, we propose to incorporate the figurative meanings from \idiomkb into idiomatic translation as a \textbf{\textit{transition}} between source idiomatic texts and target language, which is similar to Chain-of-Thought (CoT) prompting~\cite{wei2022chain}.
This method differs from earlier MT systems which directly replace idiomatic expressions with their figurative meanings~\cite{salton-etal-2014-evaluation, Modh2021UsingIF}.
As demonstrated in Figure~\ref{fig:front}, our approach effectively incorporates the figurative meaning of the idiom into the current context in the idiomatic text.

Additionally, previous automatic evaluation metrics for MT~\cite{ali-renals-2018-word,wang2023chatgpt, chen2023exploring} only analyze entire sentences without assessing idiomatic translation quality explicitly.
To address this issue, we propose an automatic evaluation metric based on GPT-4~\cite{openai2023gpt4}, which analyzes different aspects of idiomatic translation quality more effectively z

Our contributions are summarized as follows:
\begin{inparaenum}[\it 1)]
    \item To tackle the non-compositional nature of idioms, we propose a multilingual idiom knowledge base, \ie, \idiomkb, to enhance idiomatic translation, particularly for smaller LMs.
    \item We propose a better method for idiomatic translation with \idiomkb, utilizing the figurative meanings of idioms in \idiomkb as a transition for more accurate idiomatic translation.
    \item We design a new metric based on GPT-4 to assess idiomatic translation, which is better aligned with human annotations. This metric demonstrates that our approach improves idiomatic translation quality.
\end{inparaenum}

\section{Related Work}
\label{sec:related}

\paragraph{Non-compositionality and Idioms in Machine Translation}
Non-compositional multiword expressions (MWEs), notably idioms, which cannot have their meanings directly derived from their component words, present a significant challenge in various tasks~\cite{lin-1999-automatic,zhu-etal-2015-neural,hwang-hidey-2019-confirming}. 
These expressions further complicate the machine translation process due to their non-compositional nature~\cite{tzou2017does, dankers-etal-2022-transformer,dankers-etal-2022-paradox}.
Previous research has proposed specific strategies such as identifying these MWEs, particularly idioms, learning distinct embeddings for them, or reformulating them into simpler, more understandable phrases~\cite{weller-etal-2014-distinguishing,ullman-nivre-2014-paraphrasing,hashimoto-tsuruoka-2016-adaptive,constant-etal-2017-survey}.
Furthermore, evidence has shown that machine translation of idioms can be improved by incorporating parallel meanings from dictionaries or other external resources~\cite{salton-etal-2014-evaluation,zaninello-birch-2020-multiword,Modh2021UsingIF}.
Drawing inspiration from this, we build \idiomkb to assist smaller models in idiomatic translation.

\paragraph{Resources of Idiom Knowledge}
To enhance the ability of neural models to comprehend idiomatic text, researchers have developed various resources, which can be categorized into two types:
\begin{inparaenum}[\it 1)]
\item \textit{Multilingual Idioms Datasets}~\cite{moussallem-etal-2018-lidioms,agrawal-etal-2018-beating,shao-etal-2018-evaluating,10.1162/tacl_a_00572,tang2022petci}: which consist of parallel translations of idioms in one language to another to improve the idiomatic translation.
\item \textit{Monolingual Idioms Datasets}~\cite{jiang-etal-2018-chengyu,zheng-etal-2019-chid,saxena2020epie,tan-jiang-2021-learning,adewumi-etal-2022-potential}: which focus on discerning idiomatic expressions within a single language.
\end{inparaenum}
In this paper, we employ LLMs to construct \idiomkb, setting itself apart with its large scale and the feature of containing multilingual idiom meanings.

\paragraph{Large Language Models}
With the recent great success of LLMs~\cite{NEURIPS2020_1457c0d6,ouyang2022training,leiter2023chatgpt,openai2023gpt4}, \textit{in-context learning} and \textit{instruction learning} have become prevailing paradigms for deploying LLMs for downstream tasks~\cite{min-etal-2022-metaicl,ram2023context}. 
Through these paradigms, LLMs can directly generate high-quality outputs for given tasks without parameter updates~\cite{rubin-etal-2022-learning,chung2022scaling}.
For dataset construction, LLMs can be a promising alternative to resource-intensive large-scale crowd-sourcing~\cite{west-etal-2022-symbolic,yuan-etal-2023-distilling} to improve the performance of smaller and specialized models, even surpassing teacher models in some settings.
Furthermore, compared to existing traditional n-gram-based metrics like BLEU~\cite{papineni2002bleu} or neural metrics such as COMET~\cite{rei-etal-2022-cometkiwi}, LLM-based evaluation can offer greater flexibility and better alignment with humans on more challenging tasks, particularly in a reference-free setting~\cite{wang2023chatgpt, chen2023exploring, luo2023chatgpt}.
In this paper, we employ them in the construction of \idiomkb and the assessment of idiomatic translation quality.

\section{\idiomkb for Idiomatic Translation}
\label{sec:method}

In this section, we create an idiom KB, \ie, \idiomkb, to provide figurative meanings of idioms to smaller LMs to improve their translation quality.

\subsection{\idiomkb Construction: Knowledge Distillation from LLMs}\label{sec:idiomkb}
Recent research demonstrates that LLMs are promising alternatives to costly large-scale crowd-sourcing for constructing datasets~\cite{west-etal-2022-symbolic,yuan-etal-2023-distilling}.
Therefore, we leverage LLMs to distill large-scale multilingual figurative meanings of idioms to create \idiomkb.

\paragraph{Source Data Collection}
To construct a comprehensive KB, we consider the coverage of idioms in each language and gather idioms across multiple datasets to create multilingual idiom lists.
These idioms are collated from three languages, \ie, English (En), Chinese (Zh), and Japanese (Ja):
\begin{itemize}
    \item \textbf{English}: Our English idioms are sourced from the MAGPIE~\cite{haagsma-etal-2020-magpie}, IMIL~\cite{agrawal-etal-2018-beating}, EPIE~\cite{saxena2020epie}, and PIE~\cite{zhou-etal-2021-pie} datasets. MAGPIE contains Potentially Idiomatic Expressions (PIEs) in context. IMIL includes idiom translations in several Indian languages. EPIE contains idioms categorized as static or formal based on lexical changes and includes their English meanings. PIE provides parallel idiomatic and literal sentences.
    
    \item \textbf{Chinese}: For Chinese idioms, we use the PETCI~\cite{tang2022petci} and CCT~\cite{jiang-etal-2018-chengyu} datasets. PETCI includes idiom English translations from dictionaries, Google and DeepL. CCT is a cloze test dataset that includes Chengyu, the most prevalent Chinese idioms. The ChID dataset~\cite{zheng-etal-2019-chid}, another cloze-style Chinese idiom dataset, also contributes to our Chinese idiom collection.
    
    \item \textbf{Japanese}: The Japanese segment is built on the OpenMWE~\cite{hashimoto-kawahara-2008-construction} and ID10M~\cite{tedeschi-etal-2022-id10m} datasets. OpenMWE is designed for idiom identification and includes many idiomatic and literal sentences per idiom. ID10M collects idioms from several languages but does not include their meanings.
\end{itemize}

\begin{table}[t]
\small
  \centering
    \begin{tabularx}{\linewidth}{X}
    \toprule
    \rowcolor[gray]{0.95}\multicolumn{1}{c}{\textbf{KB Meaning Generation}} \\
    \makecell[l]{\color{gray}{/* \textit{Task prompt} */}\\
    Given a Chinese idiom, please write the idiom's figurative \\ \textbf{English} meaning. Please note: Idiom always expresses \\ figurative meaning which is different from literal meaning \\ of its constituent words.\\
    \color{gray}{/* \textit{Examples} */} \\
    \textbf{Case 1:} \\
    \textbf{Chinese idiom}: 明目张胆\\
    \textbf{English meaning}: straightforwardly, without any concealment\\
    ...\\
    \color{gray}{/* \textit{Test Data} */} \\
    \textbf{Case 5:} \\
    \textbf{Chinese idiom}: 一气呵成\\
    \textbf{English meaning}: \color[rgb]{0,0.39,0}\textit{to complete a task or work in one go, without } \\ \color[rgb]{0,0.39,0}\textit{stopping or taking a break}}\\
    \bottomrule
    \end{tabularx}
  \caption{Prompt for LLMs to generate \idiomkb.
  The prompt contains four examples.
  Generated texts are \color[rgb]{0,0.39,0}\textit{highlighted}.
  }
  \label{tab:kb_generation}
\end{table}

\paragraph{Idiomatic Meanings Distillation from LLMs}
We inherit the idea of symbolic knowledge distillation from models and use LLMs via in-context learning to generate figurative meanings of idioms for \idiomkb construction.
As shown in Table~\ref{tab:kb_generation}, we first manually design instructions emphasizing the non-compositional nature of idioms.
Then, we randomly select several idioms for each language pair and manually annotate their meanings from online dictionaries for reference as examples for in-context learning.
For example, we can extract the English meaning of ``一气呵成'', ``to complete a task or work in one go, without stopping or taking a break'' from the LLM output to construct \idiomkb.
By relying on the ability of LLMs to comprehend and generate accurate figurative meanings for idioms, our method is both straightforward and computationally efficient, as it avoids processing large amounts of text. 

\begin{table}[t!]
\small
\centering
\begin{tabular}{ccrc}
\toprule
\textbf{Idiom Lang.}                & \textbf{Dataset} & \textbf{Size}  & \textbf{Meaning Lang.}           \\ \midrule
\multirow{5}{*}{English (En)} & PIE     & 1,197 & En                \\
                     & IMIL    & 2,208 & Indian                \\
                     & EPIE    & 717   & En \\
                     & MAGPIE  & 1,756 & -                  \\
                     & \idiomkb    & \textbf{3,990} & En, Zh, Ja           \\ \midrule
\multirow{4}{*}{Chinese (Zh)} & PETCI   & 4,310 & En                \\
                     & CCT     & 7,395 & Zh                \\
                     & ChID    & 3,848 & -                  \\
                     & \idiomkb    & \textbf{8,643} & En, Zh, Ja            \\ \midrule
\multirow{3}{*}{Japanese (Ja)} & ID10M   & 165   & -                  \\
                     & OpenMWE & 146   & -                  \\
                     & \idiomkb    & \textbf{270}   & En, Zh, Ja       \\ \bottomrule
\end{tabular}
\caption{Comparison between \idiomkb with other existing idiom corpora in different languages.}
\label{tab:statistics}
\end{table}
We report the statistics of our KB and other existing idiom corpora for comparison.
As shown in Table~\ref{tab:statistics}, \idiomkb boasts a larger number of idioms with multilingual figurative meanings, setting it apart from the existing corpora that provide only monolingual meanings or completely lack figurative meanings.
Through the utilization of \idiomkb, we can retrieve the idiomatic meanings of idioms and effectively enhance smaller models when translating idiomatic text.

\begin{table}[t]
\small
  \centering
    \begin{tabularx}{\linewidth}{X}
    \toprule
    \rowcolor[gray]{0.95}\multicolumn{1}{c}{\textbf{I: Direct Prompt}} \\
    \makecell[l]{
    Translate the following Chinese sentence into English. \\
    \textbf{Chinese}: 为使讨论一气呵成，我们会在本报告第381至396\\段回应这些关注。\\
    \textbf{English}: \color[rgb]{0,0.39,0}\textit{We will answer these questions in the report’s 381-}\\ \color[rgb]{0,0.39,0}\textit{396 sections.}}\\
    \midrule
    \rowcolor[gray]{0.95}\multicolumn{1}{c}{\textbf{II: KB-CoT}} \\
    \makecell[l]{
     ``一气呵成'' means \color[rgb]{0.54,0.17,0.89}\textit{``to complete a task or work in one go, }\\ \color[rgb]{0.54,0.17,0.89}\textit{without stopping or taking a break''}.\\Given the above knowledge, translate the following Chinese \\sentence into English.\\
    \textbf{Chinese}: 为使讨论一气呵成，我们会在本报告第381至396\\段回应这些关注。\\
    \textbf{English}: 
    \color[rgb]{0,0.39,0}\textit{To make the discussion flow smoothly, we will respond }\\\color[rgb]{0,0.39,0}\textit{to these questions in 381-396 sections of the report.}}\\

    \midrule
    \rowcolor[gray]{0.95}\multicolumn{1}{c}{\textbf{III: Self-CoT}} \\
    \makecell[l]{
    ``一气呵成'' means \color[rgb]{0,0.39,0}\textit{``a success that comes as a result of one's }\\ \color[rgb]{0,0.39,0}\textit{without stopping or taking a break''}.\\Given the above knowledge, translate the following Chinese \\sentence into English.\\
    \textbf{Chinese}: 为使讨论一气呵成，我们会在本报告第381至396\\段回应这些关注。\\
    \textbf{English}: 
    \color[rgb]{0,0.39,0}\textit{In order to have a successful discussion, we will }\\\color[rgb]{0,0.39,0}\textit{respond to these concerns in the 381-396 sections of this report.}}\\
    \bottomrule
    \end{tabularx}
  \caption{Instruction for InstructGPT (6.7B) to translate the idiomatic text with direct (\ie, \textbf{Direct Prompt}) and CoT prompts which utilize meanings retrieved from \idiomkb (\ie, \textbf{KB-CoT}) or self-generated meanings (\ie, \textbf{Self-CoT}).
  We present the meaning retrieved from \idiomkb in {\color[rgb]{0.54,0.17,0.89}\textit{purple}}.
  Generated texts are in \color[rgb]{0,0.39,0}\textit{green}.
  }
  \label{tab:prompt_cot}
\end{table}
\subsection{\idiomkb Application: Improving Idiomatic Translation}\label{sec:idiomcot}
To bypass non-compositional expressions in idiomatic translation, we propose to prompt smaller LMs with the figurative meanings of idioms retrieved from \idiomkb to guide the translation process.

First, we identify the target idiom within the source sentence to correlate it with an idiom in \idiomkb.
% For Chinese, since idioms do not undergo grammatical changes, we can directly employ string matching to locate the idiom.
% For English and Japanese, we utilize SpaCy to lemmatize the sentence before matching the corresponding idiom.
We adopt the idiom-sentence pairs in the source datasets rather than retrieved idioms.
We think that idiom detection has already been studied, allowing us to focus our efforts on translation tasks without idiom detection as an intermediate step.
Once the target idiom is identified, we can retrieve its corresponding multilingual meanings from \idiomkb.

Then we incorporate idiom meaning in the prompt when doing idiomatic translation.
The prompt begins by formulating a short task description introducing the machine translation task.
We provide the meaning of the idiom retrieved from \idiomkb to encourage smaller LMs to focus on comprehending the non-literal meaning of the idiom before attempting a full translation.
As shown in Table~\ref{tab:prompt_cot} (II), smaller LMs can successfully translate the Chinese idiomatic text into English by obtaining the correct figurative English meaning of the Chinese idiom ``一气呵成'' from \idiomkb as a hint (\ie, KB-CoT prompt).
Compared to directly translating in Table~\ref{tab:prompt_cot} (I), the KB-CoT prompt guides LMs to focus on understanding the non-literal meaning of idioms before translating them in context, resulting in more accurate idiomatic translations. 
Furthermore, we explore the Self-CoT method in Table~\ref{tab:prompt_cot} (III), where smaller LMs generate the meaning without \idiomkb.
However, the inaccurate meaning leads to an incorrect translation, highlighting the benefits of using the correct meaning retrieved from \idiomkb to enhance the performance of smaller LMs.

\section{Metrics for Idiomatic Translation}
\label{sec:evaluating}
In this section, we aim to establish evaluation criteria for evaluating translations of idiomatic text. 
As mentioned above, idioms are unique linguistic constructs characterized by their figurative meanings which frequently deviate from their literal ones. 
This difference renders the general metrics unsuitable for idiomatic translation evaluation, as they treat idioms equally with other compositional parts and fail to have a profound understanding of idioms.
Due to the lack of aligned data between idiomatic text and its respective translations in other languages, we need to develop a reference-free metric for evaluation.
To this end, we draw from linguistic studies to formulate a comprehensive method for determining the quality of idiomatic translations using LLMs.

\begin{table}[t]
\renewcommand\cellset{\linespread{1.0}\selectfont} 
  \centering
  \small
    \begin{tabularx}{\linewidth}{X}
    \toprule
    \makecell[l]{
    \color{gray}{/* \textit{Task prompt} */}\\
    Evaluate the idiom translation in the given Chinese \\ translation of an English sentence. Focus on the idiom's \\ figurative meaning. \\
    \color{gray}{/* \textit{Evaluation Criteria} */} \\
    1 point: Ignores, mistranslates, or only translates the literal \\ meaning of the idiom. \\
    2 points: Conveys basic figurative meaning but may lack \\ refinement or have minor imperfections. \\
    3 points: Exceptional translation, accurately conveying \\ figurative meaning, context, and cultural nuances.
    \\
    \color{gray}{/* \textit{Test Data} */} \\
    Evaluate the following translation: \\
    English sentence: <source>\\
    Idiom in the English sentence: <idiom>\\
    Chinese translation: <translation>\\
    Evaluation (score only):\color[rgb]{0,0.39,0}\textit{<score>}}\\
    \bottomrule
    \end{tabularx}
  \caption{The prompt for LLMs to evaluate idiomatic translation quality. This prompt specifies in detail the translation quality and ask the LLM to generate \color[rgb]{0,0.39,0}\textit{<score>}.
  }
  \label{tab:template_evaluation}
\end{table}

\subsection{Evaluation Criteria}
As shown in Table~\ref{tab:template_evaluation}, we design a prompt based on a 1-3 point evaluation criteria to help LLM focus on idiomatic translation quality while providing detailed guidelines for each point on the scale.
In the evaluation criteria, a 1-point score reflects poor idiom translation, 2 points indicate a basic, though imperfect understanding and 3 points represent an exceptional and accurate translation incorporating figurative meaning, context, and cultural nuances.
Then LLMs are provided with the test data and asked to generate a score-only evaluation.

\subsection{Can LLMs evaluate idiomatic translation?} \label{sec:evaluation_data}
We manually construct a small-scale evaluation set on three language pairs, \ie, Chinese-to-English (Zh$\rightarrow$En), English-to-Chinese (En$\rightarrow$Zh), and Japanese-to-English (Ja$\rightarrow$En), from source language datasets:
\begin{itemize}
    \item \textbf{English}: PIE Corpus ~\cite{zhou-etal-2021-pie}, focusing on idiomatic sentence generation and paraphrasing with 1,197 idioms and 5,170 related sentences. 
    \item \textbf{Chinese}: We extract idioms from the PETCI dataset~\cite{tang2022petci} and identify sentences containing them in the WMT22 dataset~\cite{Kocmi2022FindingsOT}.
    \item \textbf{Japanese}: OpenMWE Corpus~\cite{hashimoto-kawahara-2008-construction}, containing 67,575 sentences with 146 idioms for idiom token identification.
\end{itemize}

\begin{table}[t]
    \small
    \centering
    \resizebox{\linewidth}{!}{\begin{tabular}{llccc}
    \toprule
  \textbf{Pair} & \textbf{Metric} & \textbf{Pearson's $r$} & \textbf{Spearman's $\rho$} & \textbf{Kendall's $\tau$} \\
    \midrule
    \multirow{3}{*}{Zh$\rightarrow$En} & BLEU     & 0.0936      & 0.0660       & 0.0515      \\
    & COMET     & 0.2510      & 0.2511       & 0.1984      \\
                       & GPT-4     & \textbf{0.6939}     & \textbf{0.6923}       & \textbf{0.6375}      \\
    \multirow{3}{*}{En$\rightarrow$Zh} & BLEU     & 0.3368      & 0.3277       & 0.2484      \\
    & COMET     & 0.5367      & 0.5186       & 0.4029      \\
                       & GPT-4     & \textbf{0.7891}      & \textbf{0.7879}       & \textbf{0.7338}      \\
    \multirow{2}{*}{Ja$\rightarrow$En} & COMET     & 0.4174      & 0.4031       & 0.3198      \\
                       & GPT-4     & \textbf{0.6708}      & \textbf{0.6718 }      & \textbf{0.6174} \\
    \bottomrule
    \end{tabular}}
    \caption{Correlation metrics between human and sacreBLEU, COMET or GPT-4 evaluation in different language pairs. Note that `BLEU' stands for sacreBLEU and Ja$\rightarrow$En translation does not have sacreBLEU results because the Japanese sentences lack corresponding English references.}
    \label{tab:evaluation}
\end{table}

We first randomly select 800 data samples from each source language dataset and translate them into the target language utilizing three distinct language models: two smaller LMs, InsructGPT (6.7B)~\cite{ouyang2022training} and BLOOMZ (7.1B)~\cite{muennighoff-etal-2023-crosslingual}, and one LLM, namely InstructGPT$_\texttt{003}$ (\ie, \texttt{text-davinci-003})~\cite{ouyang2022training}, a variant of GPT-3~\cite{NEURIPS2020_1457c0d6} tuned on instructions using reinforcement learning with human feedback (RLHF).
We opt for this approach to ensure that translation generated by models of different types and sizes can all be evaluated correctly.
Then we annotate 20 sentence and translation pairs for each point across all language pairs.
We compare the evaluation results of the LLMs with the human-annotated results to determine their level of consistency by calculating Pearson's $r$~\cite{pearson1920notes}, Spearman's $\rho$~\cite{spearman1987proof}, and Kendall's $\tau$~\cite{kendall1948rank} correlations.

The results in Table~\ref{tab:evaluation} demonstrate that LLMs can serve as an evaluator for the translation quality of idiomatic expressions across different language pairs.
Conversely, sacreBLEU and CometKiwi~\cite{rei-etal-2022-cometkiwi} find it challenging to align with human evaluation.
This difficulty arises largely because they lack the capacity to comprehend the nuances and meanings of idioms, thereby failing to accurately reflect human evaluations.
We also explore using few-shot prompts and observe performance decline.
This decline can potentially be due to the distraction caused by the presence of examples.
They may cause LLMs to lose focus on the current test example, thereby leading to biases.

%\section{Experiments}
%\label{sec:experiments}

\section{Experiments}

\subsection{Experimental Settings}
\paragraph{\idiomkb Construction}
For \idiomkb construction, we choose several LLMs to generate high-quality multilingual idiom meanings via in-context learning: 
GPT-3.5 series\footnote{Note that OpenAI does not release detailed information about GPT-3.5s.}, including InstructGPT$_\texttt{003}$ ($\sim$175B), ChatGPT~\cite{openai2022chatgpt} and 
multilingual LMs, \ie,  BLOOM (176B)~\cite{scao2022bloom} and BLOOMZ (176B).
GPT-3 is an auto-regressive LLM with billions of parameters achieving strong performance on NLP tasks.
InstructGPT$_\texttt{003}$ is a variant of GPT-3~\cite{NEURIPS2020_1457c0d6} fine-tuned on instructions and code via reinforcement learning with human feedback (RLHF).
ChatGPT is a dialogue-oriented model that is built on InstructGPT with RLHF.
BLOOM is a multilingual language model, and BLOOMZ is built on BLOOM using multitask-prompted fine-tuning.

\paragraph{Idiomatic Translation}
For idiomatic translation, We choose mBART (680M)~\cite{liu-etal-2020-multilingual-denoising} as a representative of multilingual Transformer models, which is an autoregressive sequence-to-sequence model and has strong performance on machine translation.
NLLB model (1.3B, distilled)~\cite{nllbteam2022language} is a supervised MT model distilled from a 54.4B Mixture-of-Experts model NLLB-200 to improve performance on low-resource languages.
We also choose InstructGPT (6.7B), BLOOMZ (7.1B) and Alpaca~\cite{alpaca}, which are all instruction-finetuned for better performance.
We also present the results of ChatGPT and GPT-4 as the upper bound for this task.

The dataset used for idiomatic translation is the same as the one employed for generating the translation evaluation set to ensure that idiomatic texts are of high quality.
Due to budget constraints, we randomly select 500 instances from each dataset for evaluation.
To ensure a fair comparison between direct prompting and KB-CoT prompting, we employ the same task description presented in Section~\ref{sec:method} and set the temperature to 0.7 for all generations.

For evaluation, we set the temperature to 0.1, intending for less randomness.
We also adopt two additional metrics, sacreBLEU and CometKiwi, where sacreBLEU represents the n-gram-based evaluation, and CometKiwi represents the reference-free neural-based evaluation.
For Zh$\rightarrow$En, we directly use the parallel English text in WMT22 as a reference.
For En$\rightarrow$Zh, we first acquire the parallel literal English text from the PIE dataset and then use ChatGPT to translate the literal text into Chinese.
We use the ChatGPT's translation result as a reference.
For Ja$\rightarrow$En, as there is no parallel text available for Japanese idiomatic text, we only report GPT-4 evaluation score and CometKiwi since they can be conducted in a reference-free setting.

\begin{table*}[t]
\small
\centering
\begin{tabular}{lllcccccccc}
\toprule
\multirow{2}{*}{\textbf{Model}} & \multirow{2}{*}{\textbf{Size}} & \multirow{2}{*}{\textbf{Setting}} & \multicolumn{3}{c}{\textbf{Zh$\rightarrow$En}}  & \multicolumn{3}{c}{\textbf{En$\rightarrow$Zh}} & \multicolumn{2}{c}{\textbf{Ja$\rightarrow$En}}  \\
\cmidrule(lr){4-6}
\cmidrule(lr){7-9}
\cmidrule(lr){10-11}
 & & & \textbf{BLEU} & \textbf{COMET} & \textbf{GPT-4} & \textbf{BLEU} & \textbf{COMET} & \textbf{GPT-4} & \textbf{COMET} & \textbf{GPT-4}\\ 
\midrule
mBART                        & 560M                  & Direct & \textbf{30.64} & \textbf{82.35} & \textbf{2.09} & \textbf{43.74} & \textbf{75.93} & \textbf{1.69} & \textbf{74.51} & \textbf{1.48} \\
NLLB                         & 1.3B                  & Direct & 25.03          & 80.75          & 1.95          & 21.64          & 64.98          & 1.65          & 70.62          & 1.44          \\
\midrule
\multirow{2}{*}{InstructGPT} & \multirow{2}{*}{6.7B} & Direct & 14.26          & 72.73          & 1.66          & 50.20          & 62.76          & 1.50          & \textbf{68.96} & 1.34          \\
                             &                       & KB-CoT & 9.64           & 73.88          & 2.08          & 13.92          & 65.49          & 1.99          & 67.50          & \textbf{1.64} \\
\multirow{2}{*}{BLOOMZ}      & \multirow{2}{*}{7.1B} & Direct & 20.60          & \textbf{79.39} & 2.11          & 49.41          & \textbf{76.88} & 2.08          & 65.29          & 1.22          \\
                             &                       & KB-CoT & 15.40          & 77.26          & \textbf{2.21} & \textbf{50.59} & 73.78          & \textbf{2.17} & 63.29          & 1.54          \\
\multirow{2}{*}{Alpaca}      & \multirow{2}{*}{7B}   & Direct & 24.80          & 71.87          & 1.54          & 21.36          & 44.28          & 1.11          & 65.82          & 1.23          \\
                             &                       & KB-CoT & \textbf{29.66} & 72.74          & 2.12          & 5.92           & 48.09          & 1.46          & 65.71          & 1.57          \\
\midrule
\multirow{2}{*}{ChatGPT}     & \multirow{2}{*}{?B} & Direct & 25.90          & \textbf{82.57} & 2.74          & \textbf{26.89} & \textbf{79.87} & 2.62          & \textbf{77.37} & 2.52          \\
                             &                       & KB-CoT & \textbf{26.42} & 82.01          & \textbf{2.82}  & 24.87          & 77.91          & \textbf{2.71} & 76.32          & \textbf{2.61}     \\
\multirow{2}{*}{GPT-4}     & \multirow{2}{*}{?B} & Direct & 21.40          & \textbf{82.43} & 2.73          & 24.87 & \textbf{79.74} & 2.73          & \textbf{77.55} & 2.63          \\
                             &                       & KB-CoT & \textbf{26.42} & 81.38          & \textbf{2.86}  & \textbf{32.21}          & 77.89          & \textbf{2.83} & 76.02          & \textbf{2.69}     \\
  \bottomrule
\end{tabular}
\caption{The translation performance of LMs in different language pairs.
The source language is either directly translated (Direct) or generated via KB-CoT prompting with meaning from \idiomkb(KB-CoT). The best results are \textbf{bolded}. Note that `BLEU' stands for sacreBLEU and Ja$\rightarrow$En translation does not have sacreBLEU results because the Japanese sentences lack corresponding English references.
}
\label{tab:main_result}
\end{table*}

\begin{table}[t]
\small
\resizebox{\linewidth}{!}{\begin{tabular}{lclcc}
\toprule
\textbf{Model}                                                      & \textbf{Setting}              & \textbf{Resource} & \textbf{Zh$\rightarrow$En} & \textbf{Ja$\rightarrow$En} \\ \midrule
\multirow{6}{*}{InstructGPT (6.7B)} & Direct               & -                          & 1.66 & 1.34          \\ \cmidrule(l){2-5} 
     & \multirow{5}{*}{KB-CoT} & Self                       & 1.69 & 1.35          \\
     &                      & BLOOM                      & 1.97 & 1.36         \\
     &                      & BLOOMZ                     & 2.07 & 1.47          \\
     &                      & InstructGPT$_\texttt{003}$ & 2.07 & 1.46          \\
     &                      & ChatGPT                    & \textbf{2.08} & \textbf{1.64} \\ \midrule
\multirow{6}{*}{BLOOMZ (7.1B)}                                        & Direct               & -                          & 2.11 & 1.22          \\ \cmidrule(l){2-5} 
     & \multirow{5}{*}{KB-CoT} & Self                       & 2.14 & 1.22          \\
     &                      & BLOOM                      & 2.15 & 1.28          \\
     &                      & BLOOMZ                     & 2.20  & 1.41          \\
     &                      & InstructGPT$_\texttt{003}$ & 2.19 & 1.38          \\
     &                      & ChatGPT                    & \textbf{2.21} & \textbf{1.54} \\ \bottomrule\end{tabular}}
\caption{Translation results with different sourced meanings retrieved from \idiomkb constructed by different LLMs.}
\label{tab:curie}
\end{table}

\subsection{Can \idiomkb Improve Idiomatic Translation for Small LMs?}

\paragraph{Main Results and Analysis}\label{sec:main_results}
The results in Table~\ref{tab:main_result} show that:
\begin{inparaenum}[\it 1)]
    \item KB-CoT prompting with meaning retrieved from \idiomkb consistently surpasses direct prompting for smaller models, revealing the universality of our KB-CoT method;
    \item For direct prompting, BLOOMZ (7.1B) outperforms InstructGPT (6.7B) and Alpaca while gaining less for KB-CoT prompting. This means that BLOOMZ (7.1B) itself shows relatively strong performance on idiomatic translation. However, its ability to combine instruction with translation is relatively weaker;
    \item For ChatGPT and GPT-4, KB-CoT prompting also improves idiomatic translation quality compared to direct prompting. This indicates that the use of KB-CoT prompting is not solely beneficial to smaller LMs, but can also enhance the performance of LLMs such as ChatGPT and GPT-4 for idiomatic translation task. However, it should be noted that GPT-4 may exhibit a bias towards the translation generated by GPT models~\cite{liu-etal-2023-g};
    \item Both pre-trained model mBART and supervised-MT model NLLB encounter difficulties in accurately translating idiomatic text;
    \item scareBLEU does not align well with the idiomatic translation quality score assessed by GPT-4, indicating that n-gram based metrics are unsuitable for idiomatic translation evaluation. CometKiwi, as a sentence-level metric, performs better than sacreBLEU but cannot evaluate idiom translation quality properly.
\end{inparaenum}

\begin{figure}[t]
    \centering
    \includegraphics[width=0.9\linewidth]{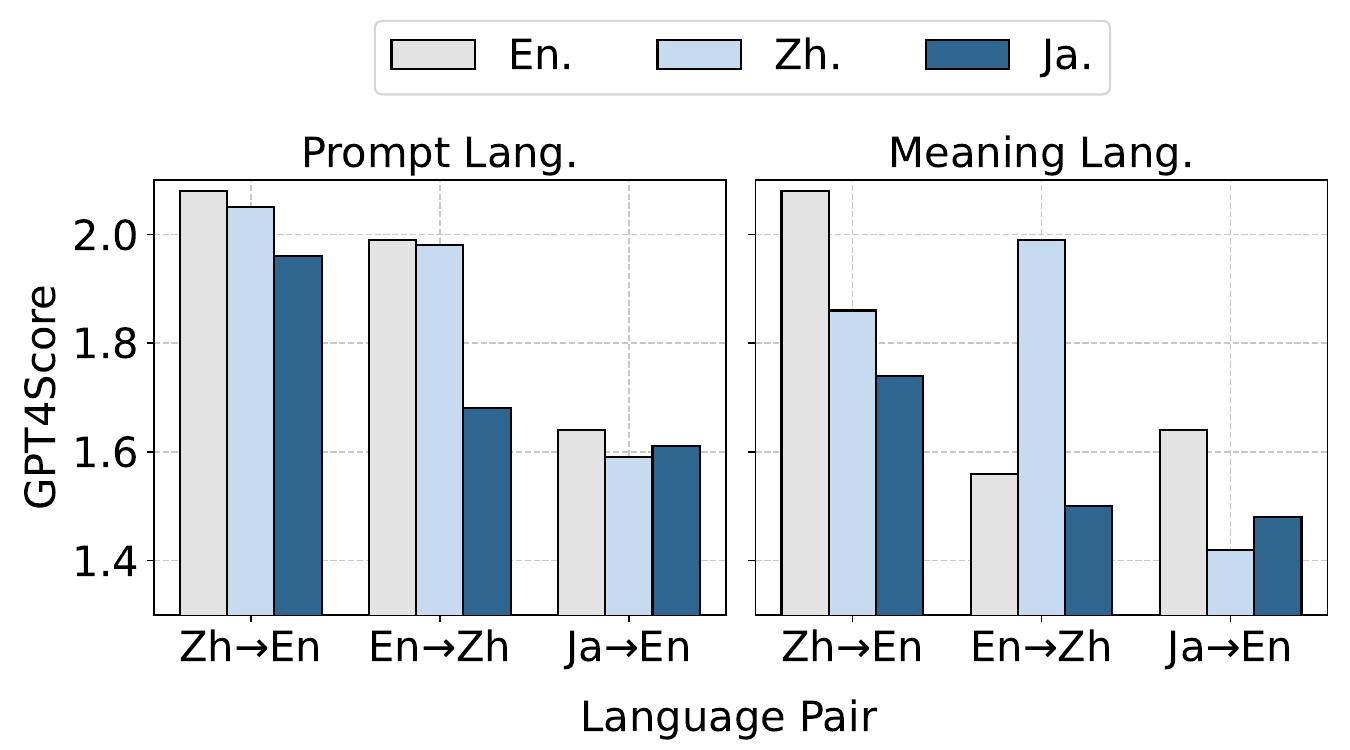}
    \caption{The performance of InstructGPT (6.7B) + KB-CoT, with prompts and meanings of different languages generated by ChatGPT.}
    \label{fig:language}
\end{figure}

\paragraph{What language should we use to form prompts?}
We compare the performance of InstructGPT (6.7B) with KB-CoT prompts of different languages.
The results in Figure~\ref{fig:language} demonstrate the superiority of utilizing English prompts.
A potential reason is that the instruction-tuning datasets predominantly consist of English data.
The findings suggest English is ideal for creating KB-CoT prompts across various language pairs, which could stem from the inherent complexity and rich vocabulary of the language, enabling precision in delivering complex instructions~\cite{shi2023language}.

\paragraph{What language should we use for idiom meaning?}
To construct \idiomkb, we ask LLMs to generate multilingual meanings for idioms.
Since this translation task involves multiple languages, deciding which language meaning to provide the model as a reference is critical.
We compare InstructGPT (6.7B)'s performance using KB-CoT prompts in English and evaluate its ability to incorporate meanings in different languages on the same idiomatic translation datasets as above.
The results in Figure~\ref{fig:language} indicate that when performing Zh$\rightarrow$En,  En$\rightarrow$Zh and Ja$\rightarrow$En translations, utilizing the meaning in the target language yields better results.
This could be due to the inherent structural and semantic differences between these languages.
When translating, retaining the meaning in the target language ensures better context understanding and cultural sensitivity.
Moreover, syntax variances, for example, word order in Chinese or Japanese differs greatly from in English, making direct translation complex and often inaccurate. 
Hence the utilization of meaning in the target language proves more effective.

\subsection{How is the quality of \idiomkb under human evaluation?}

\begin{figure}[t]
    \centering
    \includegraphics[width=0.8\linewidth]{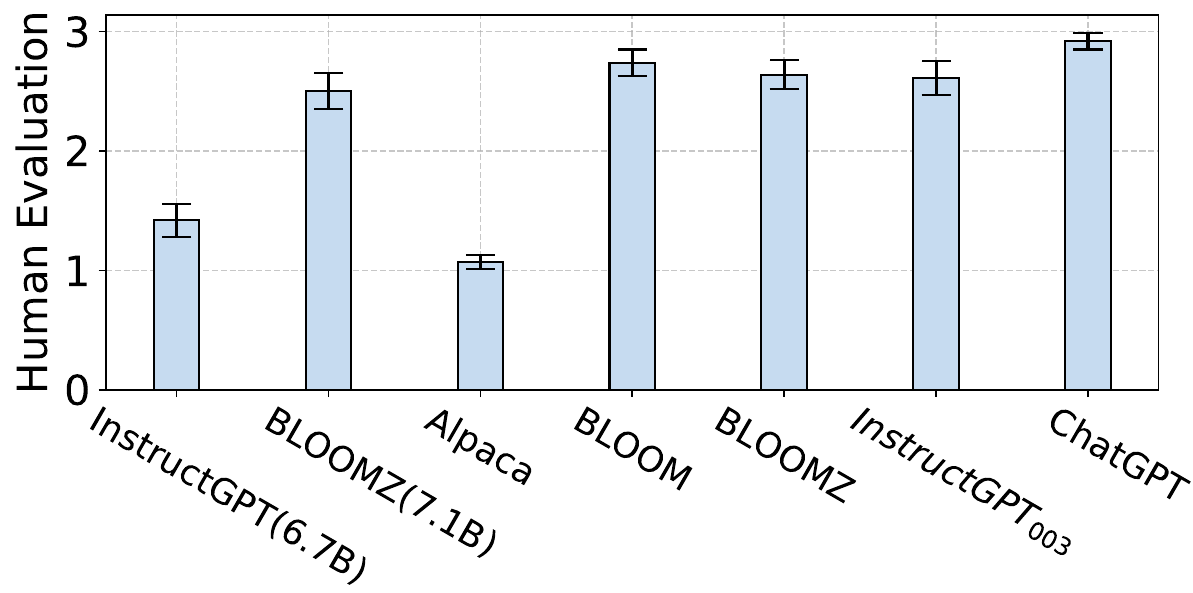}
    \caption{Human evaluation of different meaning sources. We use a wide range of models, including LLMs and relatively smaller models to obtain meanings.}
    \label{fig:meaning_human_eval}
\end{figure}

\begin{figure}[t]
    \centering
    \includegraphics[width=0.9\linewidth]{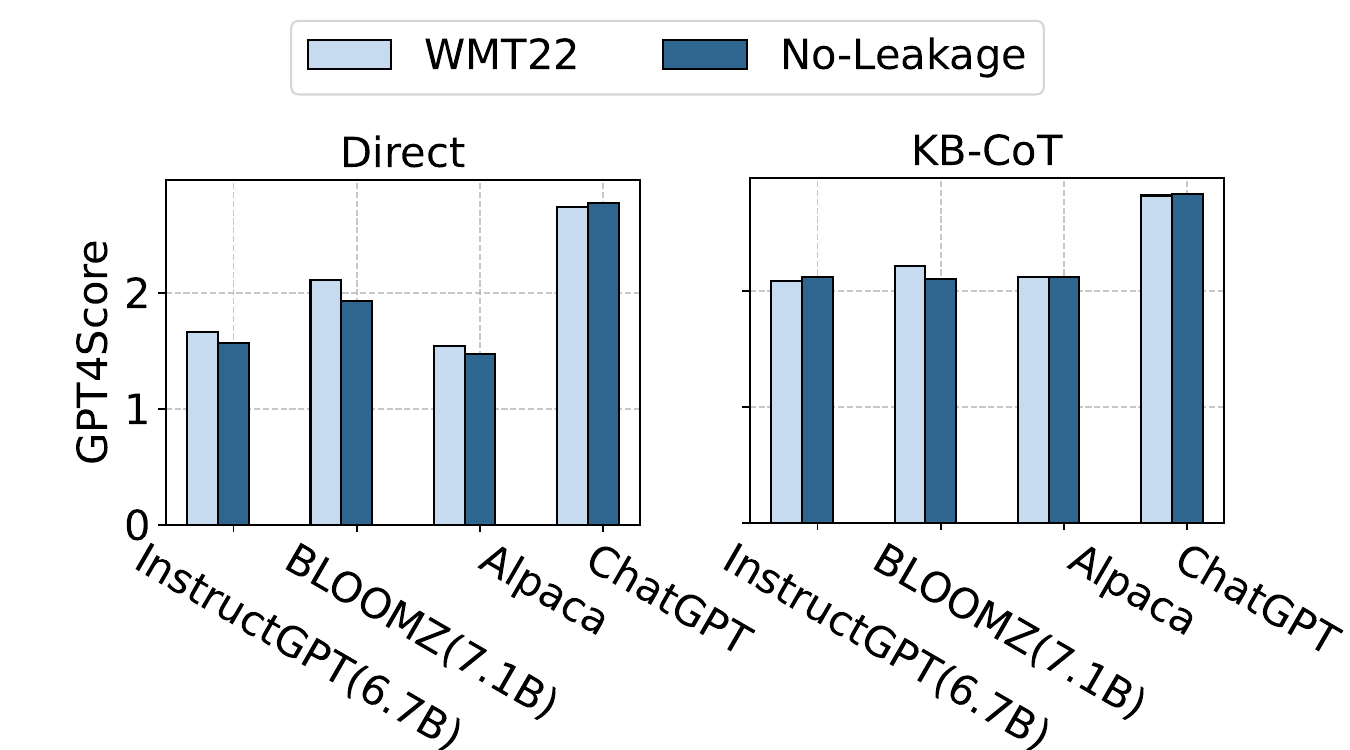}
    \caption{Model performance on idiomatic text extracted from WMT22 and no-leakage dataset (Zh$\rightarrow$En).}
    \label{fig:leakage}
\end{figure}

\paragraph{Which model generates the best \idiomkb?}
We manually annotate the quality of idiom meaning generated by different LMs utilizing the method in $\mathsection$~\ref{sec:idiomkb}.
Each model generates Chinese meanings of 100 randomly selected Chinese idioms.
To evaluate idiom quality, we assign points ranging from 1-3, with 1 indicating a completely inaccurate meaning, 2 indicating the meaning requiring minor refinements, and 3 indicating a perfect capture of nuanced cultural meanings.
The results in Figure~\ref{fig:meaning_human_eval} show that \idiomkb generated by ChatGPT produces the highest quality results, which are consistent with the translation performance presented in Table~\ref{tab:curie}.
In contrast, smaller models struggle to generate high-quality idiom meanings independently.

\paragraph{Will data leakage affect idiomatic translation?}
Although we strive to select the most current datasets, there remains a possibility that LMs may have encountered the specific sentences we employ. 
To address this concern, we follow the approach of \citet{zhu2023multilingual} and manually create a no-leakage dataset consisting of 60 recently published news sentences that contain idioms.
Then we compare the Zh-En idiomatic translation performance of LMs on our idiomatic translation test set extracted from WMT22 and the no-leakage dataset.
The results in Figure~\ref{fig:leakage} indicate that the effect of data leakage on idiomatic translation is negligible.

\begin{table}[!t]
\small
  \centering
    \begin{tabularx}{\linewidth}{X}
    \toprule
    \makecell[l]{
    \textbf{Source sentence}: ``即使是发达经济国家也不能\\永远{\color[rgb]{0.54,0.17,0.89}{\textbf{寅吃卯粮}}}。''\\
    \textbf{Figurative meaning}: A metaphor for economic hardship \\ and inadequate income, borrowing and misappropriating in \\ advance \\
    \textbf{Literal meaning}: Eat the grain of the year of the Tiger \\ during the year of the Rabbit. \\
    \textbf{Reference}: ``Even developed economies cannot {\color[rgb]{0.12,0.56,1}{[live beyond}} \\ {\color[rgb]{0.12,0.56,1}{their means forever]}}. ''}\\
    \midrule
    \makecell[l]{
    \textbf{mBART}: ``Even the advanced economies will not be able to\\ {\color[rgb]{0.70,0.13,0.13}{<eat their fill>}} forever. '' {$\Rightarrow$ \underline{Literal Translation Error}}\\
    \textbf{NLLB}: ``Even in developed economies, \color[rgb]{0.70,0.13,0.13}{<food cannot be eaten>} \\forever. '' {$\Rightarrow$ \underline{Literal Translation Error}}}\\
    \midrule
    \makecell[l]{
    \textbf{InstructGPT (6.7B) (Direct)}: \\``Even a developed country cannot indefinitely \color[rgb]{0.70,0.13,0.13}{<subsist on rice } \\ {\color[rgb]{0.70,0.13,0.13}{porridge>}}. ''{$\Rightarrow$ \underline{Literal Translation Error}}\\
    \textbf{InstructGPT (6.7B) (KB-CoT)}: \\``Even a developed economy can't always indefinitely {\color[rgb]{0.12,0.56,1}{ [live off}} \\ {\color[rgb]{0.12,0.56,1}{tomorrow's food]}} indefinitely. ''{$\Rightarrow$ \underline{Correct Translation}}}\\
    \midrule
    \makecell[l]{
    \textbf{BLOOMZ (7.1B) (Direct)}:\\ ``Even in developed economies, it is not possible to {\color[rgb]{0.70,0.13,0.13}{<live on}} \\ {\color[rgb]{0.70,0.13,0.13}{borrowed time>}} forever. ''{$\Rightarrow$ \underline{Idiom Misunderstanding}}\\
    \textbf{BLOOMZ (7.1B) (KB-CoT)}:\\``Even in developed economies, it is not possible to {\color[rgb]{0.12,0.56,1}{[live off}} \\ {\color[rgb]{0.12,0.56,1}{tomorrow's resources today]}}. ''{$\Rightarrow$ \underline{Correct Translation}}}\\
    \bottomrule
    \end{tabularx}
  \caption{A comparative case study illustrating the performance of mBART, NLLB, InstructGPT (6.7B), and BLOOMZ (7.1B) on Zh-En idiomatic translation. Each model's translation of the Chinese idiom {\color[rgb]{0.54,0.17,0.89}{``寅吃卯粮''}} is given in {\color[rgb]{0.70,0.13,0.13}{<red>}} (inaccurate) or {\color[rgb]{0.12,0.56,1}{[blue]}} (accurate). }
  
  \label{tab:case_study}
\end{table}
\paragraph{Case Study}
We present the case study in Table~\ref{tab:case_study}. 
``寅吃卯粮'' is a Chinese idiom referring to spending resources in advance, which literally translates as eating the food stored up for the next year.
``Yin'' (寅) and ``Mao'' (卯), which refer to the year of tiger and rabbit in this context, are the 3rd and 4th terms in the 12 Earthly Branches widely used in traditional Chinese calendars and horoscopic astrology, and thus the model needs to understand the cultural nuances to translate this idiom accurately.
Consistent with the main experiment, all four translation models struggle with this idiom to differing extents.
mBART and NLLB generate literal translations such as ``eat their fill'' and ``food cannot be eaten'', respectively.
Similarly, InstructGPT (6.7B) with direct prompting also makes a literal translation error, mistakenly translating the idiom as ``subsist on rice porridge'' while BLOOMZ (7.1B) misunderstands the idiom as ``live on borrowed time''.
Conversely, when enhanced by \idiomkb and KB-CoT, the performance of InstructGPT (6.7B) and BLOOMZ (7.1B) significantly improves. 
Both models successfully capture and convey the figurative meaning of the idiom.
Interestingly, there are idioms of similar meanings in different cultures and languages, such as, in this case, the English idiom ``robbing Peter to pay Paul'', which means using next month's (or period's) resources to cover this month's (or period's) expenses.
We believe our \idiomkb would be very useful in studying the cross-culture alignments of idioms, which we leave for future work.

\section{Conclusion}
\label{sec:conclusion}
In this paper, we present a solution to tackle the challenges of idiomatic translation, which can be applied to various sizes of models.
We develop \idiomkb, a multilingual idiom knowledge base that leverages the figurative meanings of idioms as a transitional aid to prevent non-compositional expressions. 
We build \idiomkb from LLMs, which are finite and can be stored offline, and then retrieve the idiom meanings from the KB and add them in the CoT prompting to improve translation quality.
Furthermore, we introduce an automatic evaluation method with GPT-4 to assess the translation quality of idioms, showing the effectiveness of our approach. 
We believe \idiomkb will be a valuable resource to advance the research on idiomatic translation and the study of the cross-culture alignment of idiomatic expressions.

\section*{Acknowledgement}

We thank the anonymous reviewers for their valuable feedback.
This work is supported by Science and Technology Commission of Shanghai Municipality Grant (No. 22511105902), National Natural Science Foundation of China (No.62102095), and Shanghai Municipal Science and Technology Major Project (No.2021SHZDZX0103).

% \clearpage
\bibliography{aaai24}
\end{CJK}
\end{document}